\pdfoutput=1
\documentclass{article}
\usepackage[dvipsnames,table]{xcolor}
\usepackage{graphicx}
\usepackage[preprint]{neurips_2024}
\usepackage{algorithm}
\usepackage{algorithmic}
\usepackage{graphicx}           
\usepackage[utf8]{inputenc}     
\usepackage[T1]{fontenc}        
\usepackage{hyperref}           
\usepackage{url}                
\usepackage{booktabs}           
\usepackage{amsfonts}           
\usepackage{nicefrac}           
\usepackage{microtype}          
\usepackage[utf8]{inputenc}
\usepackage[T1]{fontenc}
\usepackage{amsmath}
\definecolor{best}{HTML}{fdf4bb}
\definecolor{secondbest}{HTML}{bffdbb}

\title{Small Contributions, Small Networks:

Efficient Neural Network Pruning Based on Relative Importance}

\author{
  Mostafa Hussien \\
  ÉTS, University of Quebec, Canada\\
  \texttt{mostafa.hussien@etsmtl.ca} \\
  \AND
  Mahmoud Afifi \\
  Google \\
  \AND
  Kim Khoa Nguyen\\
  ÉTS, University of Quebec, Canada \\
   \AND
   Mohamed Cheriet \\
  ÉTS, University of Quebec, Canada\\
}

\begin{document}

\maketitle

\begin{abstract}
Recent advancements have scaled neural networks to unprecedented sizes, achieving remarkable performance across a wide range of tasks. However, deploying these large-scale models on resource-constrained devices poses significant challenges due to substantial storage and computational requirements. Neural network pruning has emerged as an effective technique to mitigate these limitations by reducing model size and complexity. In this paper, we introduce an intuitive and interpretable pruning method based on activation statistics, rooted in information theory and statistical analysis. Our approach leverages the statistical properties of neuron activations to identify and remove weights with minimal contributions to neuron outputs. Specifically, we build a distribution of weight contributions across the dataset and utilize its parameters to guide the pruning process. Furthermore, we propose a \textit{Pruning-aware Training} strategy that incorporates an additional regularization term to enhance the effectiveness of our pruning method. Extensive experiments on multiple datasets and network architectures demonstrate that our method consistently outperforms several baseline and state-of-the-art pruning techniques.
\end{abstract}

\section{Introduction}
\label{sec:intro}

Deep learning has achieved remarkable results across various fields, from computer vision to natural language processing, by generating highly effective models like large language models (LLMs) (e.g., \cite{GPT, LLaMA, gemini}), which have demonstrated significant improvements in multiple applications. These models have shown significant improvements in a wide range of applications, including machine translation (\cite{lewis2019bart}), question answering (\cite{raffel2020exploring}), and image classification (\cite{abdelhamed2024you}). However, as deep neural networks (DNNs) grow in size to handle increasingly complex problems, they require immense computational resources, both in terms of memory and processing power.

Network pruning, also referred to as network or model compression, aims to reduce the size of these networks, thereby decreasing their computational costs. This is achieved by removing specific weights from the model, setting them to zero based on certain pruning criteria. DNN pruning methods can be categorized into different groups based on the nature of the approach (e.g., data-free versus data-driven, or based on the pruning criteria used). We refer the reader to \cite{survey} for a thorough discussion of these categories. From a high-level perspective, we can categorize pruning methods into structural pruning (e.g., \cite{wang2020pruning, huang2018data, liu2017learning, theus2024towards, ganjdanesh2024jointly, shi2024towards, gadhikar2024masks, wu2024auto, guo2023automatic, fang2023depgraph, he2017channel}), where entire filters or channels are removed, and unstructured pruning (e.g., \cite{tanaka2020pruning, mason2024makes, choi2023sparse, lee2018snip, su2020sanity, wang2020picking, bai2022dual, mocanu2018scalable, magnitude, wanda}), which performs weight-wise pruning. In the latter case, the network is typically retrained after pruning and it is common for pruning to be performed iteratively, where a smaller set of weights is selected for removal (i.e., set to zero), followed by retraining or fine-tuning the pruned model. This process is repeated until the target final pruning ratio is reached.

While data can provide valuable insights into how each neuron (or node) contributes to the final result, the majority of unstructured pruning methods rely solely on neuron weights, focusing on defining criteria to measure the significance of individual weight values. For instance, the magnitude-based pruning metric \cite{magnitude} removes weights by eliminating those with magnitudes below a certain threshold.

Recent work, such as Wanda \cite{wanda}, enhances the traditional weight magnitude pruning metric by incorporating input activations. Designed specifically for LLMs, Wanda is based on the observation that, at a certain scale, a small subset of hidden state features exhibits significantly larger magnitudes than others \cite{dettmers2022gpt3}. The pruning score in Wanda is computed as the product of the weight magnitude and the norm of the corresponding input activations, recognizing that input features can vary considerably in the scale of their output features. While Wanda demonstrates promising results, it does not fully capture the true contribution of each neuron weight to the output feature, given the input features.

In this paper, we introduce a data-driven, unstructured pruning method that utilizes training data—or a subset thereof—to approximate the distribution of each weight's importance in the network based on its contribution to the output of its corresponding node. By applying the \textit{Central Limit Theorem}, we model the aggregated importance of weights as a normal distribution, which enables us to estimate the mutual information between a weight and the output of its associated node. This mutual information quantifies how much knowing the weight reduces uncertainty about the node's output. Consequently, the more sensitive the node's output is to changes in a weight, the more important that weight is and the less likely it is to be pruned. The gradient of the activation function has a clear impact on the performance of the pruning method, as they affect the distribution of the node's output. Our proposed method is firmly grounded in both statistical analysis and information theory, drawing connections to the Central Limit Theorem and mutual information. Preliminary experiments demonstrate that our method consistently yields more accurate models, even at high compression rates, compared to alternative approaches.
\section{Method}
\label{sec:method}
\subsection{Activation Blind Range}

\begin{figure}
    \centering
    \includegraphics[width=1.0\linewidth]{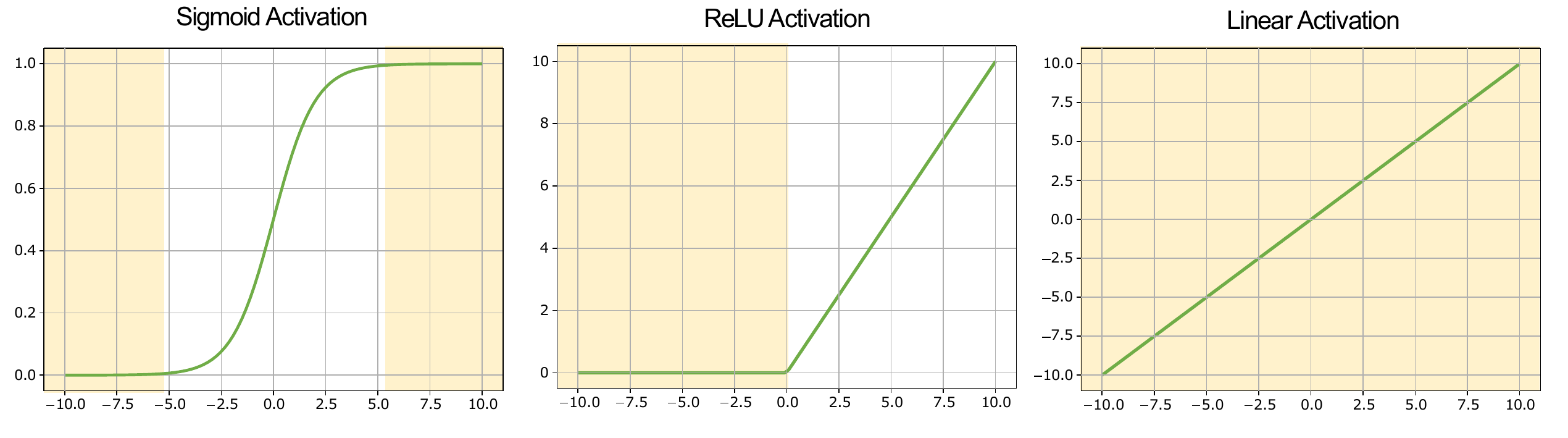}
    \caption{The blind range of various activation functions, defined as the interval in which the gradient of the activation function is zero. In this range, the function's output remains constant, providing a ``safe zone'' for pruning, where changes to the weights do not affect the model’s output. A wider blind range offers greater flexibility for pruning algorithms, allowing for more aggressive weight reduction without impacting performance. The blind range is highlighted by a yellow color in this figure.}
    \label{fig:blindRange}
\end{figure}
The role of nonlinear activation functions has been widely studied in various aspects of neural network architectures, including their impact on training convergence, weight initialization, and stability. For example, activation functions play a crucial role in gradient propagation, influencing issues such as the vanishing and exploding gradient problems, which are critical for training deep networks. However, less focus has been given to the impact of activation functions on the susceptibility of neural networks to pruning. This study explores how the choice of activation function impacts the extent to which an architecture can be pruned without causing significant degradation in accuracy. 

We introduce the concept of the \textit{``Blind Range''} for a typical activation function, which refers to the interval where the derivative of the activation function is zero, see Fig. \ref{fig:blindRange}. In other words, the blind range represents the input range over which the activation function’s output remains constant. For instance, in the case of the \textit{ReLU} activation, this range spans from negative infinity to zero.

We propose that the blind range of activation functions provides a safe zone for pruning, where if pruning a weight causes the activation output to fall within this range, the output of the corresponding node remains unchanged, and as a result, the overall model performance is preserved. Additionally, small deviations from this blind range can be efficiently corrected during the fine-tuning phase. However, the effect of pruning may vary across different data points. To address this variability, it is necessary to adopt a statistical approach. Specifically, we suggest empirically constructing a distribution to quantify the impact of each weight across different subsets of the dataset, enabling more informed and robust pruning decisions. This is explained in more details in the next sections.

\subsection{Relative Weight Contributions}
\begin{figure}
    \centering
    \includegraphics[width=0.85\linewidth]{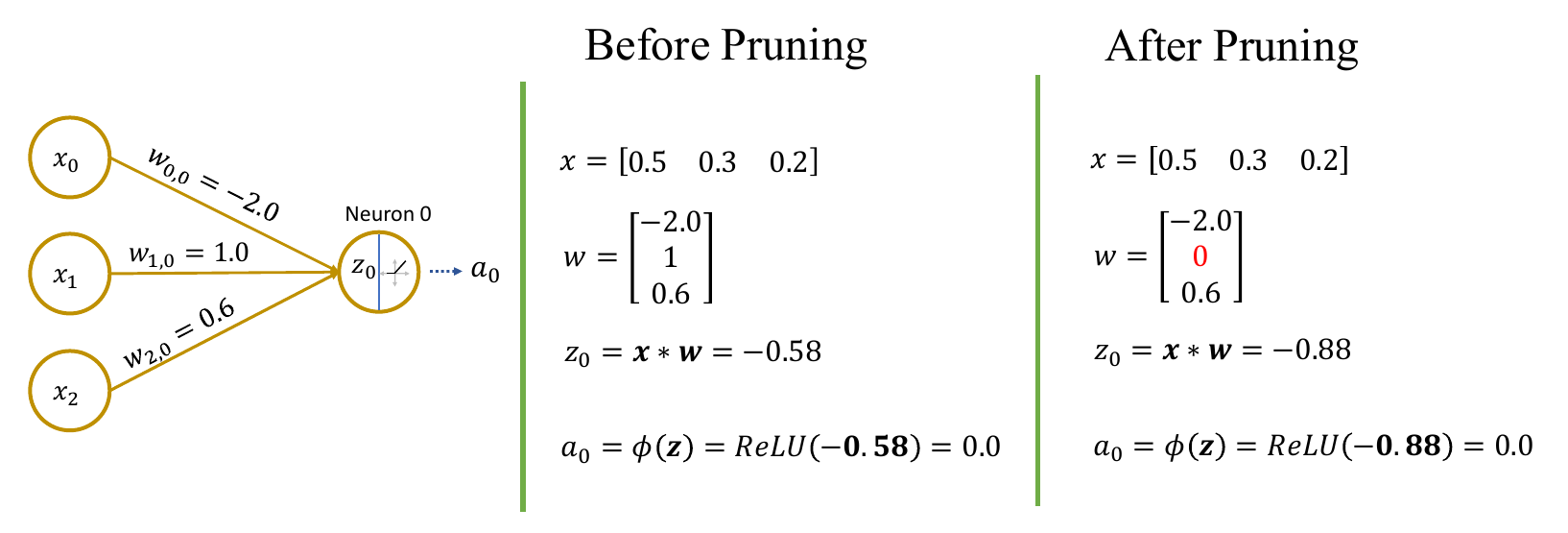}
    \caption{A simplified illustrative example of the proposed pruning method applied for a single-node architecture. The left panel depicts a single node receiving three inputs, each connected by a corresponding weight. The center panel shows the calculation of the node output, \( a_n \), prior to any pruning. The right panel demonstrates the effect of pruning the second weight, \( w_{1,0} = 1.0 \), and its subsequent impact on the final node output.}
    \label{fig:NumericalExample}
\end{figure}
Figure \ref{fig:NumericalExample} illustrates how a node contributes to the activation of its associated neuron within the neural network architecture. Specifically, we analyze the contribution of a weight $w_{i,j}$ in a layer that receives an input vector of size $I$ and adopts an activation function $f$. We define the contribution function $\varsigma(\cdot)$ of a weight $w_{i,j}$ as:
\begin{equation}
\label{eq:contribution}
\begin{split}
    a_j = f(\sum_{n=1}^{I} x_n \times w_{(n,j)}), \\
    \bar{a_j} = f(\sum_{n=1}^{I} x_n \times w_{(n,j)}) \;\;\;  n\neq i\\
    \varsigma (w_{n,j}) = \left| \left(a_j -\bar{a_j}\right)  / a_j\right|,
\end{split}
\end{equation}

where $x_n$ is the $n^{th}$ input of the layer, $w_{n,j}$ is the weight connecting the $n^{th}$-input to the $j^{th}$-node in a typical layer. The magnitude of this contribution determines the actual importance of the corresponding weight in the final node activations and, consequently, indicates the extent to which the weight can be pruned. Given that the contributions of each weight vary with different data points, and considering the large number of data points, the distribution of these contributions over the epochs approaches a Gaussian distribution according to the \textit{Central Limit Theorem} \cite{sirignano2020mean}. Utilizing the first-order statistics of the contributions' distribution, we define a weight function that assigns a scalar value to each weight, representing its importance, as shown in Equation \ref{eq:importanceFunc}.
\begin{equation}
  \mathbb{I}(w_{i,j}) = s \times [\alpha \times \mathbb{E}(\varsigma (w_{i,j})) + \beta \times \frac{1}{\epsilon + \sigma (\varsigma (w_{i,j}))}],
\label{eq:importanceFunc}
\end{equation}

\noindent where $s = 2^i$ is a decaying factor that controls the contribution of each layer, $\alpha$, $\beta$, weight parameters to control the importance of the mean and the standard deviation, respectively. The term $\epsilon$ is a small number to avoid division by zero. After calculating the importance value of each weight based on its contribution, the pruning becomes a straight forward process by applying iterative weight pruning given by Algorithms. \ref{alg:pruning}.

\begin{figure}
    \centering
    \includegraphics[width=\linewidth]{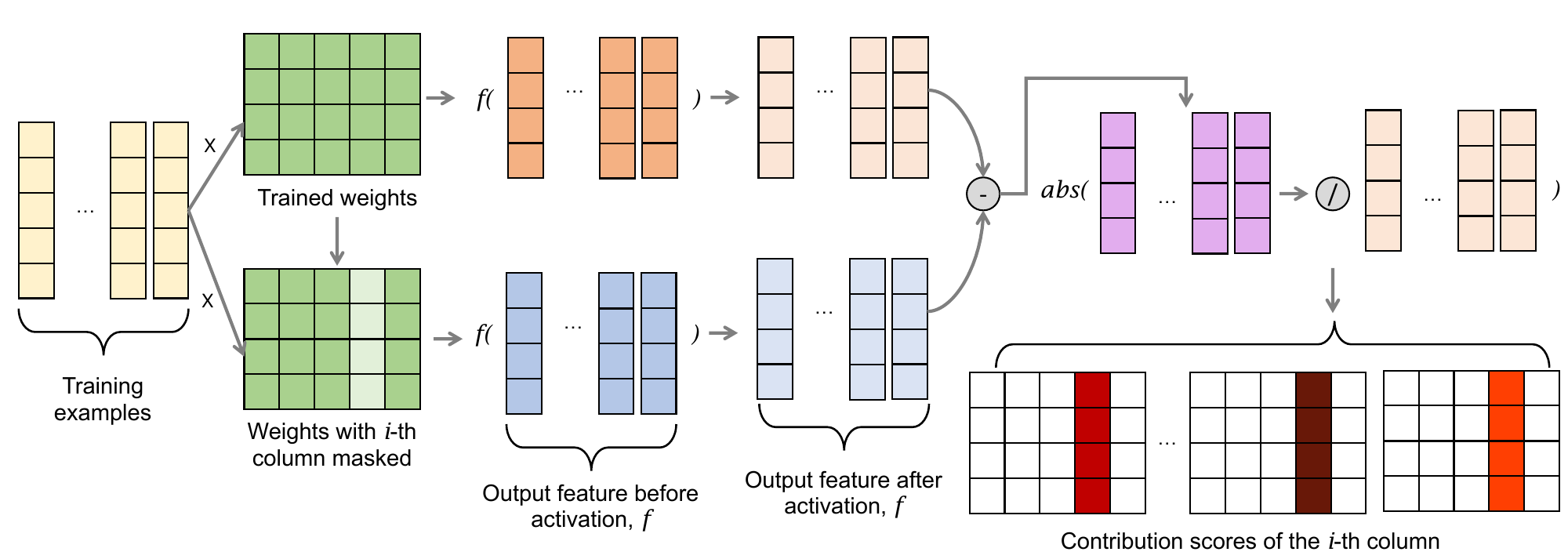}
    \caption{We propose a pruning metric based on the relative weight contribution of each neuron. The contribution function is computed by measuring the relative contribution of each neuron in every network layer. The illustration shows the $i$-th column in the fully connected weight matrix. We feed training samples into this layer to compute the output features (in blue). Then, we mask the $i$-th column (set to zero) and compute the output without its influence. The relative contribution of the $i$-th weights (in red) is then computed for each training example, representing the distribution of neuron contributions of this column. 
    \label{fig:main}}
\end{figure}

\begin{algorithm}
\caption{Iterative Weight Pruning Algorithm}
\label{alg:pruning}
\begin{algorithmic}[1]
\REQUIRE Trained model $\text{Model}$ consisting of $L$ layers; Dataset $\mathcal{D}$; Target pruning percentage $P_t$; Pruning per iteration $P_i$
\ENSURE Pruned model $\text{Model}_p$

\STATE $\text{Model}_p \leftarrow \text{Model}$
\WHILE{$P_t > 0$}
    \IF{$P_t \leq P_i$}
        \STATE $P_{\text{current}} \leftarrow P_t$
        \STATE $P_t \leftarrow 0$
    \ELSE
        \STATE $P_{\text{current}} \leftarrow P_i$
        \STATE $P_t \leftarrow P_t - P_i$
    \ENDIF
    \FORALL{weights $w$ in $\text{Model}_p$}
        \STATE Compute $\mathbb{I}(w)$
    \ENDFOR
    \STATE $\text{threshold}$ = $F^{-1}_{\mathbb{I}}(P_{\text{current}})$ \; //The $P_{\text{current}}$-th percentile of the weight importance distribution
    \FORALL{weights $w$ in $\text{Model}_p$}
        \IF{$\mathbb{I}(w) < \text{threshold}$}
            \STATE Set $w \leftarrow 0$
        \ENDIF
    \ENDFOR
    \STATE Retrain $\text{Model}_p$ on dataset $\mathcal{D}$
\ENDWHILE
\RETURN $\text{Model}_p$
\end{algorithmic}
\end{algorithm}

\subsection{Pruning-aware Training}
In this section, we introduce a novel regularization term for training that enhances the effectiveness of our pruning algorithm. As previously discussed, the probability of pruning a weight increases as its mutual information with the node's output decreases. In other words, if a node's output frequently falls within the blind range, the weights associated with that node are more likely to be pruned. Therefore, increasing the probability of a neuron's output being in the blind range improves the pruning process.

To achieve this, we propose a loss function that incorporates an additional term penalizing the model based on the magnitude of the nodes' outputs. The modified loss function is defined as:
\begin{equation}
\label{eq:proposedLoss}
    \mathcal{L}_{n} = \mathcal{L}_{\text{o}} + \lambda_{rL1} \sum_{i} \left| a_i \right|  
\end{equation}

where \(\mathcal{L}_{\text{n}}\) is the new loss function used for training the model, \(\mathcal{L}_{\text{o}}\) is the original loss function, \(\lambda_{rL1}\) is a regularization hyperparameter that controls the contribution of the regularization term to the final loss, and \(a_i\) is the output of the \(i\)-th neuron. Note that the regularization component, which is the $L1$ norm of the nodes output, encourages the dispersion of neuron outputs. As demonstrated in the results section, incorporating this regularization term significantly enhances the performance of our proposed pruning algorithm. The following section presents the theoretical foundations of the proposed pruning technique in era of \textit{Information Theory}.

\textbf{Speedup:} To enhance the efficiency of the proposed algorithm, we replaced the weight-by-weight computation of weight contributions with an optimized matrix multiplication algorithm that computes the contributions of entire columns of weights simultaneously. This modification resulted in a significant improvement in computational speed compared to the one-by-one method (see Fig.\ref{fig:main}).

\section{Mutual Information}
In this section, we analyze the mutual information between the activation output of the $i^{th}$-node in the $l^{th}$-layer, $a^l_i$ and the weight $w^l_{ij}$ connecting neuron $i$ to the $j^{th}$ input \cite{czyz2024beyond}. As known, the activation output is given by:
\begin{equation}
    a^l_i = \phi(z_i), \text{where}\; z_i=\sum_j w_{ij} x_j,
\end{equation}
with $\phi(\cdot)$ representing the activation function and $x_j$ is the $j^{th}$-input from the preceding layer. The mutual information $I(a_i; w_{ij})$ quantifies the reduction in uncertainty of $a_i$ due to knowledge of $w_{ij}$, calculated as:
\begin{equation}
    I(a_i; w_{ij}) = H(a_i) - H(a_i \mid w_{ij}).    
\end{equation}
Here, $H(a_i)$ is the entropy of the activation output $a_i$, defined as $H(a_i) = - \sum_{a_i} P(a_i) \log P(a_i)$, and $H(a_i \mid w_{ij})$ is the conditional entropy given by:
\begin{equation}
    H(a_i \mid w_{ij}) = - \int_{w_{ij}} P(w_{ij}) \sum_{a_i} P(a_i \mid w_{ij}) \log P(a_i \mid w_{ij}) \, dw_{ij}. 
\end{equation}

A high mutual information indicates a significant dependency between $a_i$ and $w_{ij}$, implying that pruning $w_{ij}$ would substantially alter $a_i$ and potentially degrade network performance. Conversely, a low mutual information suggests that $w_{ij}$ has little influence on $a_i$, making it a suitable candidate for pruning without affecting the model's accuracy. Incorporating the concept of the Blind Range—where the activation function's derivative $\phi'(z_i) = 0$ and $a_i$ remains constant—we observe that mutual information $I(a_i; w_{ij})$ is minimal when $a_i$ operates predominantly within this range. This reinforces our strategy to prune weights with large likelihood of being associated with activations in the Blind Range. To compute $I(a_i; w_{ij})$ in practice, we estimate the probability distributions $P(a_i)$ and $P(a_i \mid w_{ij})$ by collecting samples of activation outputs and corresponding weights across the dataset. For continuous variables, we compute differential entropy:
\begin{equation}
h(a_i) = - \int P(a_i) \log P(a_i) \, da_i,
\end{equation}

\begin{equation}
    h(a_i \mid w_{ij}) = - \int P(w_{ij}) \int P(a_i \mid w_{ij}) \log P(a_i \mid w_{ij}) \, da_i \, dw_{ij},
\end{equation}
 By calculating these entropy values and the resulting mutual information, we can rank the weights based on their $I(a_i; w_{ij})$ values. We then define a threshold $\tau$ below which weights are considered for pruning. This method allows us to systematically identify and remove redundant weights, enhancing model efficiency while maintaining performance. This analysis of mutual information provides a robust, mathematically grounded framework for neural network pruning based on the concept of the activation's \textit{Blind Range} and contributing to the development of more efficient and compact models.

\section{Experimental Results}
\label{sec:results}

In this section, we validate our method on the MNIST dataset \cite{mnist} for the image classification task. We fixed the network architecture to a simple fully connected layer network consisting of 3 layers. The first layer has 784 $\times$ 392 weight neurons, accepting flattened images of size 28 $\times$ 28, and outputs 392 features, followed by a ReLU activation function. The second layer processes the latent representation and produces 196 output features, again followed by a ReLU activation function. The final layer outputs a 10-dimensional vector representing the logits for the MNIST classes. We utilized cross-entropy loss with a learning rate of 0.001 for 10 epochs and optimized the network using the Adam optimizer with betas = (0.9, 0.999).

To perform pruning, we adopted an iterative pruning strategy. After training the original model, we iteratively prune the network as follows: First, we compute a score for each weight neuron based on the pruning metric, either using our method's contribution metric or the criteria of other methods for comparison. Next, we determine a threshold value from these scores, based on a fixed pruning ratio per iteration, and create a mask that zeros out neurons with scores below the threshold. This process is repeated iteratively, pruning neurons at each step until the final target pruning ratio is reached. After each pruning iteration, we fine-tune the model by applying the current pruning mask and training for one epoch using a learning rate of 0.0001. The mask is updated in each iteration to include the newly pruned neurons, representing all pruned neurons by the end of the process.

Note that since our method relies on input data to compute the contribution of each neuron, we investigate the impact of using different subsets of the training dataset for this purpose in the pruning process. The final mean and standard deviation of the computed contributions are calculated across the training examples used in our pruning procedure to derive the final contribution score, as described in Equation \ref{eq:importanceFunc}. We apply the same approach for examining different subsets of training data for Wanda \cite{wanda}, as it also depends on input data to compute the pruning metric for weights.

Table \ref{tab:ablation} shows the results of using different values of $\beta$ in Equation \ref{eq:importanceFunc} and the impact of the layer decay factor, $s$. The best results were achieved with $\beta = 1e-7$ and incorporating the layer decay factor, $s$, which intuitively makes sense, as the importance of neurons should be scaled based on the layer depth in the network.

We consider the following pruning metrics for comparisons: (1) random pruning, where neurons are randomly selected for pruning based on the target ratio; (2) magnitude-based pruning \cite{magnitude}, where neurons with the smallest weight magnitudes are pruned; and (3) Wanda \cite{wanda}, which extends magnitude-based pruning by additionally considering the input feature norm statistics. The results of these comparisons are shown in Table \ref{tab:comp1}, where we report performance with different pruning ratios per iteration (and thus different numbers of pruning iterations), as well as using varying portions of the training dataset for data-driven methods (i.e., Wanda \cite{wanda} and ours). As shown, our method achieves the best results, even when using only 2\% of the training data to compute the contribution score. Our method consistently outperforms other methods across all settings.

\begin{table}[t]
\centering
\caption{\textbf{Ablation studies} on the impact of $\beta$ and the layer decay factor, $s$, on our results are presented. The experiments were conducted using the ReLU activation function with a simple fully connected network, which achieved 97.02\% accuracy without pruning. In these experiments, the target pruning ratio was set to 50\% (i.e., eliminating 50\% of the network weights). All reported results are presented as percentages. The best results are highlighted in \colorbox{best}{yellow}. \label{tab:ablation}}
\scalebox{0.85}{
\begin{tabular}{l|cccc|}
 &
  \multicolumn{4}{c|}{\textbf{Pruning ratio per iteration}} \\ \cline{2-5} 
\multicolumn{1}{|l|}{\textbf{Configuration}} &
  \multicolumn{1}{c}{25\%} &
  \multicolumn{1}{c}{15\%} &
  \multicolumn{1}{c}{10\%} &
  \multicolumn{1}{c|}{5\%} \\ \hline
\multicolumn{1}{|l|}{$\alpha$ = 1, $\beta$ = 1e-7, w/o $s$} &
  97.61 & 97.45 & 97.91 & 97.93 \\
\multicolumn{1}{|l|}{$\alpha$ = 1, $\beta$ = 0, w/ $s$} & 97.62 & 97.49 & 97.90 & 98.0 \\
\multicolumn{1}{|l|}{$\alpha$ = 1, $\beta$ = 1e-5, w/ $s$} & 96.14 & 96.63 & 97.28 & 97.59 \\
\multicolumn{1}{|l|}{$\alpha$ = 1, $\beta$ = 1e-7, w/ $s$} & \colorbox{best}{97.84} & \colorbox{best}{97.52} & \colorbox{best}{97.92} & \colorbox{best}{98.03}
\end{tabular}}
\end{table}

\begin{table}[t]
\centering

\caption{\textbf{Comparisons with other methods.} The original model achieved an accuracy of 97.02\% without pruning, and the target pruning ratio was set to 50\% (i.e., eliminating 50\% of the network weights). We compare our method with random pruning, magnitude pruning \cite{magnitude}, and Wanda \cite{wanda}. For the data-driven methods (Wanda and ours), we evaluated performance with varying percentages of training data, noted alongside each method's name. All reported results are presented as percentages. The best results are highlighted in \colorbox{best}{yellow}, while the second-best results are highlighted in \colorbox{secondbest}{green}. \label{tab:comp1}}
\scalebox{0.85}{
\begin{tabular}{l|cccc|}
                                      & \multicolumn{4}{c|}{\textbf{Pruning ratio per iteration}} \\ \cline{2-5} 
\multicolumn{1}{|l|}{\textbf{Method}} & 25\%          & 15\%         & 10\%         & 5\%         \\ \hline
\multicolumn{1}{|l|}{Random} & 93.19 &  91.24   &  79.22 &   9.80  \\
\multicolumn{1}{|l|}{Magnitude \cite{magnitude}}       &  97.29  & 97.77 & 97.99 & 98.11 \\
\multicolumn{1}{|l|}{Wanda (0.5\%) \cite{wanda}} & 97.47 & 97.72 & 98.16 & 98.3 \\
\multicolumn{1}{|l|}{Wanda (2\%) \cite{wanda}}     &  97.44 & 97.82 & 98.18 & 98.32  \\
\multicolumn{1}{|l|}{Wanda (10\%) \cite{wanda}}    & 97.4 & 97.87 & 98.21 & 98.33 \\
\multicolumn{1}{|l|}{Wanda (20\%) \cite{wanda}}    & 97.38 & 97.86 & 98.2 & 98.28  \\
\multicolumn{1}{|l|}{Wanda (50\%) \cite{wanda}}   & 97.41 & 97.91 & 97.92 & 98.26 \\
\multicolumn{1}{|l|}{Wanda (100\%) \cite{wanda}}   & 97.42 & 97.96 & 97.98 & 98.31  \\
\multicolumn{1}{|l|}{Ours (0.5\%)}    &  97.84 & 97.52 & 97.92 & 98.03 \\
\multicolumn{1}{|l|}{Ours (2\%)}      &  98.04 & 98.11 & 98.37 & 98.33 \\
\multicolumn{1}{|l|}{Ours (10\%)}     &  98.26 & \colorbox{best}{\textbf{98.36}} & 98.41 & 98.34  \\
\multicolumn{1}{|l|}{Ours (20\%)}     & 98.24 & 98.35 & \colorbox{best}{\textbf{98.47}} & \colorbox{secondbest}{98.41}  \\
\multicolumn{1}{|l|}{Ours (50\%)}     &  \colorbox{best}{\textbf{98.36}} & \colorbox{secondbest}{98.34} & \colorbox{secondbest}{98.38} & 98.40 \\
\multicolumn{1}{|l|}{Ours (100\%)}    & \colorbox{secondbest}{98.29} & \colorbox{best}{\textbf{98.36}} & 98.30 & \colorbox{best}{\textbf{98.44}}        
\end{tabular}}
\end{table}

Table \ref{tab:diff-activation} shows the results of experiments using activation functions other than ReLU, which was used in previous experiments. To evaluate the generalizability of our approach, we report results using Leaky ReLU, Sigmoid, and Tanh activation functions. As shown, our method continues to outperform other pruning techniques across all tested activation functions.

\begin{table}[t]
\centering
\caption{\textbf{Results using different activation functions} with accuracy without pruning indicated alongside each activation function. In these experiments, the final pruning ratio was set to 50\% (i.e., eliminating 50\% of the network weights). We compare our method with random pruning, magnitude pruning \cite{magnitude}, and Wanda \cite{wanda}. For data-driven methods (Wanda and ours), we evaluated the performance with varying percentages of training data, indicated alongside each method's name. All reported results are presented as percentages. The best results are highlighted in \colorbox{best}{yellow}, while the second-best results are highlighted in \colorbox{secondbest}{green}. \label{tab:diff-activation}}
\scalebox{0.68}{
\begin{tabular}{l|cccccccccccc|}
              & \multicolumn{12}{c|}{\textbf{Pruning ratio per iteration}}                  \\ \cline{2-13} 
                & \multicolumn{4}{c|}{\textbf{Leaky ReLU (97.35\%)}} & \multicolumn{4}{c|}{\textbf{Sigmoid (97.64\%)}} & \multicolumn{4}{c|}{\textbf{Tanh (96.12\%)}} \\ \hline
\textbf{Method} & 25\%   & 15\%  & 10\%  & \multicolumn{1}{c|}{5\%}  & 25\%  & 15\%  & 10\% & \multicolumn{1}{c|}{5\%} & 25\%      & 15\%      & 10\%      & 5\%      \\ \hline
Random & 93.19 & 92.69 & 88.09 & \multicolumn{1}{c|}{9.80} &  93.94 &	83.55 & 84.70 & \multicolumn{1}{c|}{9.80} &  93.36 & 92.21 & 39.73 & 9.80 \\
Magnitude   \cite{magnitude}  &  97.15 & 97.58 & 97.89 & \multicolumn{1}{c|}{97.87} &  89.45 & 92.28 & 94.26  & \multicolumn{1}{c|}{96.09} &  92.74 & 94.14 & 95.66 & 97.20\\
Wanda (0.5\%) \cite{wanda}&  96.77 & 97.21 & 97.78 & \multicolumn{1}{c|}{97.88} &  91.19 & 94.66 & 96.22 & \multicolumn{1}{c|}{96.91} &  93.11 & 94.9 & 96.47 & 97.35  \\
Wanda (20\%) \cite{wanda} &  96.68 & 97.15 & 97.72 & \multicolumn{1}{c|}{97.86} &  91.37 & 94.44 & 95.97 & \multicolumn{1}{c|}{96.89} &  93.3 & 94.68 & 96.39 & 97.20 \\
Wanda (100\%) \cite{wanda}&  96.63 & 97.20 & 97.79 & \multicolumn{1}{c|}{97.89} &  91.40 & 94.42 & 96.18 & \multicolumn{1}{c|}{96.83} &  93.21 & 94.67 & 96.33 & 97.26\\
Ours (0.5\%)  &  97.03 & 97.5 & 97.81 & \multicolumn{1}{c|}{98.01} &  94.64 & 95.44 & 96.09 & \multicolumn{1}{c|}{96.62 } & 96.17 & 96.36 & 96.86 & 97.1  \\
Ours (20\%)   &  \colorbox{secondbest}{97.97} & \colorbox{best}{\textbf{98.18}} & \colorbox{secondbest}{98.12} & \multicolumn{1}{c|}{\colorbox{best}{\textbf{98.22}}} &  \colorbox{secondbest}{95.79} & \colorbox{secondbest}{96.47} & \colorbox{secondbest}{97.14} & \multicolumn{1}{c|}{\colorbox{secondbest}{97.56}} &  \colorbox{best}{\textbf{97.09}} & \colorbox{secondbest}{97.35} & \colorbox{secondbest}{97.51} & \colorbox{secondbest}{97.76} \\
Ours (100\%)  &  \colorbox{best}{\textbf{98.01}} & \colorbox{secondbest}{98.12} & \colorbox{best}{\textbf{98.21}} & \multicolumn{1}{c|}{\colorbox{secondbest}{98.18}} &  \colorbox{best}{\textbf{96.08}} & \colorbox{best}{\textbf{96.66}} & \colorbox{best}{\textbf{97.15}} & \multicolumn{1}{c|}{\colorbox{best}{\textbf{97.70}}} &  \colorbox{secondbest}{97.08} & \colorbox{best}{\textbf{97.50}}	& \colorbox{best}{\textbf{97.54}} & \colorbox{best}{\textbf{97.82}}
\end{tabular}
}
\end{table}

So far, we have fixed the final target pruning ratio at  50\% of the original neurons. In Table \ref{tab:diff-pruning-ratios},we show the results for various final pruning ratios. In this experiment, the pruning ratio per iteration was set to 5\%. The results indicate that even with an aggressive pruning ratio of 75\% (reducing the original network size), our method achieves an accuracy of 97.63\%, which is better than with the original model accuracy of 97.02\%. In contrast, other methods experience higher reductions in accuracy, with magnitude-based pruning \cite{magnitude} yielding 92.52\% and Wanda \cite{wanda} achieving 90.32\%.

\begin{table}[t]
\centering
\caption{\textbf{Results using different target pruning ratio} (10\%, 50\%, 75\%). Here, we used a fixed pruning ratio per iteration (5\%) and compare our method with magnitude pruning \cite{magnitude} and Wanda \cite{wanda}. For data-driven methods (Wanda and ours), we evaluated the performance by using the full training data for pruning. All reported results are presented as percentages. The best results are highlighted in \colorbox{best}{yellow}, while the second-best results are highlighted in \colorbox{secondbest}{green}. \label{tab:diff-pruning-ratios}}
\scalebox{0.8}{
\begin{tabular}{l|ccc|}
                                    & \multicolumn{3}{c|}{\textbf{Target pruning ratio}} \\ \cline{2-4} 
\multicolumn{1}{|l|}{\textbf{Method}}        & 10\%   & 50\%   & 75\%   \\ \hline
\multicolumn{1}{|l|}{Magnitude \cite{magnitude}}     &  98.58 & 98.11 & 92.52 \\
\multicolumn{1}{|l|}{Wanda \cite{wanda}} &  \colorbox{secondbest}{98.60} & \colorbox{secondbest}{98.31} & \colorbox{secondbest}{90.32} \\ 
\multicolumn{1}{|l|}{Ours}  &   \colorbox{best}{\textbf{98.66}} & \colorbox{best}{\textbf{98.44}} & \colorbox{best}{\textbf{97.63}}
\end{tabular}
}
\end{table}

\begin{table}[t]
\centering
\caption{\textbf{Results with $\lambda_{r_{L1}}$ using different activation functions}. For each activation function, the accuracy of original model (without pruning) with and without $\lambda_{r_{L1}}$ are as follows: ReLU (w/o $\lambda_{r_{L1}}$: 97.02\%, w/ $\lambda_{r_{L1}}$: 97.4\%), Tanh (w/o $\lambda_{r_{L1}}$: 96.12\%  , w/ $\lambda_{r_{L1}}$:  95.78 \%), and Sigmoid (w/o $\lambda_{r_{L1}}$:  97.64\%  , w/ $\lambda_{r_{L1}}$:    ). The target pruning ratios were set to 50\% and 75\%, with a 25\% pruning ratio per iteration. We compare our method with random pruning, magnitude pruning \cite{magnitude}, and Wanda \cite{wanda}. For data-driven methods (Wanda and ours), we evaluated the performance with varying percentages of training data, indicated alongside each method's name. All reported results are presented as percentages. The best results are highlighted in \colorbox{best}{yellow}, while the second-best results are highlighted in \colorbox{secondbest}{green}. \label{tab:lambda-regularizer-l1}}
\scalebox{0.8}{
\begin{tabular}{l|ccc|ccc|}& \multicolumn{3}{c|}{\textbf{Target pruning ratio (50\%)}} & \multicolumn{3}{c|}{\textbf{Target pruning ratio (75\%)}} \\ \cline{2-7} 
\multicolumn{1}{|l|}{\textbf{Method}} &
  \textbf{ReLU} &
  \textbf{Tanh} &
  \textbf{Sigmoid} &
  \multicolumn{1}{c}{\textbf{ReLU}} &
  \multicolumn{1}{c}{\textbf{Tanh}} &
  \multicolumn{1}{c|}{\textbf{Sigmoid}} \\ \hline
\multicolumn{1}{|l|}{Random }       &   93.34  &  92.92  &  81.94 &     84.71  &  83.97 &  68.75 \\
\multicolumn{1}{|l|}{Magnitude \cite{magnitude}}    &  96.43   & 94.87   &  93.54    &   68.1   &   74.05 &   60.23   \\
\multicolumn{1}{|l|}{Wanda (0.5\%) \cite{wanda}} &  97.16 & 95.34  &   96.02  &   90.83  &   91.07  &  85.46 \\
\multicolumn{1}{|l|}{Wanda (100\%) \cite{wanda}} &  97.21  &  95.48 &  96.07 &  90.74  &   91.07   &  84.84 \\ 
\multicolumn{1}{|l|}{Ours (0.5\%)}  &  \colorbox{secondbest}{97.80}  &   \colorbox{best}{\textbf{97.10}}   &   \colorbox{secondbest}{97.39} & \colorbox{secondbest}{96.52}  & \colorbox{best}{\textbf{95.02}} &    \colorbox{secondbest}{92.52} \\
\multicolumn{1}{|l|}{Ours (100\%)}  &  \colorbox{best}{\textbf{97.87}}  &  \colorbox{secondbest}{96.98} &  \colorbox{best}{\textbf{97.51}}  & \colorbox{best}{\textbf{98.15}}  &  \colorbox{secondbest}{94.79}   & \colorbox{best}{\textbf{92.62}}
\end{tabular}}
\end{table}

Table \ref{tab:lambda-regularizer-l1}  demonstrates the results of applying the proposed loss function in Equation \ref{eq:proposedLoss} with the indicated regularization term. As discussed in Section \ref{sec:method}, the hyperparameter $\lambda_{r_{L1}}$ controls the contribution of the regularization to the overall loss function. By utilizing this regularization term, we achieved a 75\% reduction in model size without significant loss in accuracy, as evidenced by the data presented in Table \ref{tab:lambda-regularizer-l1}.

\section{Conclusion}
In this paper, we have introduced an interpretable, simple, yet effective pruning method. By defining the concept of the activation blind range, we investigated the underexplored aspect of how activation functions affect an architecture's susceptibility to pruning. We presented a statistical framework for the proposed pruning method, grounded in the \textit{Central Limit Theorem} and \textit{Mutual Information} concepts. Our findings conclude that, for unstructured pruning, considering the mutual information between each weight and its associated node leads to a simple and powerful pruning strategy. Moreover, considerable improvements have been obtained by leveraging what we named as \textit{'Pruning-aware Training'} that incorporates an extra term that encourage the model to push the nodes output toward the blind range of the activations. The experimental results confirm the effectiveness of the proposed method across various experimental settings.

\bibliographystyle{unsrtnat}
\bibliography{ref}





\end{document}